\documentclass[conference]{IEEEtran}
\IEEEoverridecommandlockouts
\usepackage{cite}
\usepackage{amsmath,amssymb,amsfonts}
\usepackage{algorithm}
\usepackage{algpseudocode}
\usepackage{graphicx}
\usepackage{multirow}
\usepackage{textcomp}
\usepackage{xcolor}
\def\BibTeX{{\rm B\kern-.05em{\sc i\kern-.025em b}\kern-.08em
    T\kern-.1667em\lower.7ex\hbox{E}\kern-.125emX}}
\begin{document}

\title{Federated Object Detection for Quality Inspection in Shared Production\\
\thanks{This project was funded and done in collaboration with Huawei Technologies Düsseldorf GmbH, at the European Research Center
in Munich.}
}

\author{\IEEEauthorblockN{1\textsuperscript{st} Vinit Hegiste}
\IEEEauthorblockA{\textit{Chair of Machine Tools and Control Systems} \\
\textit{RPTU Kaiserslautern-Landau}\\
Kaiserslautern, Germany \\
vinit.hegiste@rptu.de}
\and
\IEEEauthorblockN{2\textsuperscript{nd} Tatjana Legler}
\IEEEauthorblockA{\textit{Chair of Machine Tools and Control Systems} \\
\textit{RPTU Kaiserslautern-Landau}\\
Kaiserslautern, Germany \\
tatjana.legler@rptu.de}
\and
\IEEEauthorblockN{3\textsuperscript{rd} Kirill Fridman}
\IEEEauthorblockA{\textit{Huawei Technologies Düsseldorf GmbH}\\
Düsseldorf, Germany \\
kirill.fridman@huawei.com}
\and
\IEEEauthorblockN{4\textsuperscript{th} Martin Ruskowski}
\IEEEauthorblockA{\textit{Innovative Factory Systems (IFS)} \\
\textit{German Research Center for Artificial Intelligence (DFKI)}\\
Kaiserslautern, Germany \\
martin.ruskowski@dfki.de}
}

\maketitle

\begin{abstract}
Federated learning (FL) has emerged as a promising approach for training machine learning models on decentralized data without compromising data privacy.
In this paper, we propose a FL algorithm for object detection in quality inspection tasks using YOLOv5 as the object detection algorithm and Federated Averaging (FedAvg) as the FL algorithm.
We apply this approach to a manufacturing use-case where multiple factories/clients contribute data for training a global object detection model while preserving data privacy on a non-IID dataset. 
Our experiments demonstrate that our FL approach achieves better generalization performance on the overall clients' test dataset and generates improved bounding boxes around the objects compared to models trained using local clients' datasets.
This work showcases the potential of FL for quality inspection tasks in the manufacturing industry and provides valuable insights into the performance and feasibility of utilizing YOLOv5 and FedAvg for federated object detection.
\end{abstract}


\section{Introduction}
Object detection (OD) is a pivotal deep learning task, sparked by breakthroughs like YOLO (You Only Look Once) \cite{Redmon.2016} and SSD (Single Shot Detector) \cite{Liu.2016}, which unify object classification and localization into a single step, enhancing efficiency and speed. While Faster R-CNN \cite{Ren.2016}, Mask R-CNN \cite{He.2017}, EfficientDet \cite{Tan.2020} adopt two-stage detectors, they are surpassed by the speed and accuracy of single shot detectors like YOLO and SSD. In manufacturing, OD proves indispensable, facilitating solutions for quality inspection, error classification, object tracking, and more.
Despite common applications among companies such as quality inspection and fault detection, data sharing is hindered by privacy and competitive concerns. The persistent challenge for custom object detectors remains dataset annotation. Here, horizontal federated learning (FL) offers a solution. FL effectively addresses privacy in collaborative learning scenarios by enabling joint model training without centralized data sharing \cite{McMahan.2017}. Applying horizontal FL to OD enhances the global federated model by aggregating class-specific samples from diverse clients. This approach augments model robustness compared to local training.

In our previous work, we applied FL to image classification tasks in manufacturing for quality inspection, showing that this approach can achieve comparable performance to centralized learning while preserving data privacy \cite{Hegiste.2022}. In some cases, the global model could even generalize to the dataset's feature space better than the model trained using centralized learning.
This paper extends our previous work by proposing a FL algorithm for object detection in quality inspection tasks, using YOLOv5 (You Look Only Once version 5) \cite{Jocher.2020} as the object detector algorithm and federated averaging (FedAvg) \cite{McMahan.2017} as the FL algorithm. OD is a critical task in quality inspection, where detecting and localizing defects in images is essential for ensuring product quality. By using a FL approach, we can train a global object detection model that incorporates data from multiple factories while maintaining data privacy. Our experiments demonstrate the feasibility and effectiveness of this approach for quality inspection tasks in manufacturing.
In addition, we would like to highlight how the quality inspection service based on FL can be integrated into the Production Level 4 Shared Production ecosystem based on Skill-Based Production. To support the requirements of Shared Production, we will offer the quality inspection service as Software-as-a-service on a marketplace. That means that the quality inspection service should have the ability of self-description. To achieve interoperability and provide the software to a marketplace, we use the Asset Administration Shell (AAS) to describe the software service in the form of submodels.

\section{Related work}
For this paper, we used YOLOv5 \cite{Jocher.2020} as this version of YOLO has been there for the last three years, while being updated and therefore stable as compared to the latest YOLOv7 \cite{Wang.2020} and other YOLO versions.
The output from OD can also be used for various purposes like locating objects in a fixed environment, counting objects, object segmentation, detecting and classifying faults, etc. These utilities hold promise in industrial quality inspection, motivating collaborative training of a global model to detect diverse errors.
Federated object detection (FedOD) finds relevance here, allowing the global model to learn from local data features across clients.
There are very few papers published in the field of FedOD, with no detailed research in applying the algorithm for specific custom use cases in manufacturing setting. 
Luo et al. try to tackle this problem by introducing the algorithm to real life dataset in \cite{Luo.2021}, but the application just focuses on images from surveillance cameras for general object detection on common objects from the streets, and does not tackle any specific use case or neither provides a depth analysis regarding the precision of the global model or any comparison with the local client models. It also uses YOLOv3, which is easily outperformed by the current state-of-the-art YOLOv5. 
The rest of the papers using FL with YOLOv5 algorithm such as, \cite{Bommel.2021} introduces Active learning during FedOD to solve the unlabeled data problem. \cite{Zhang.2023}  introduces FedVisionBC, a blockchain-based federated learning system for visual object detection, to solve the privacy challenges of FL.
FedCV \cite{He.2021} focuses on creating a framework for automating the process of FedOD, but the repository is still a work in progress and difficult to use for custom use cases, as the paper demonstrates the model for basic public datasets.
Su et al. \cite{Su.2022}, uses RetinaNet \cite{Lin.2017} with backbone ResNet50 \cite{HeKaiming.2016} as their detection algorithm- \cite{Su.2022} uses their custom algorithm which uses an ensemble step to tackle the challenges arising from cross-domain FedOD.
Our paper follows a similar algorithm of FedOD using FedAvg\cite{McMahan.2017}, and tries to explore the usability of Federated OD for quality inspection in shared production scenario with non-IID dataset using  different use cases.

The concept of the Asset Administration Shell (AAS) serves as the implementation of the digital twin in the context of Industry 4.0 \cite{Tasnim.2022}.
Acting as a digital representation of an asset or service, the AAS comprises various sub-models that encompass all relevant information and functionalities of the asset or service. This includes its features, characteristics, properties, and capabilities. The AAS facilitates communication through diverse channels and applications, serving as the crucial link between physical objects and the connected, digital, and distributed world.
Initially employed for creating digital representations of physical assets, the AAS can now also accommodate software modules. Its integration is a fundamental prerequisite for incorporating production modules into emerging production architectures.
Looking ahead to 2025, the flexible production network aims to operate seamlessly through the employment of the digital platform Gaia-X \cite{Alexopoulos.2023}.
For assets such as machines and services to be integrated into the European data platform Gaia-X in the future, they must meet specific technical standards, including considerations related to security and skill descriptions. This relevant information is encapsulated within the AAS, ensuring compliance and enabling assets to participate effectively within the Gaia-X ecosystem.

\section{Methodology} \label{methodology}

\begin{figure}[h]
    \centering
    \includegraphics[width= 0.48\textwidth]{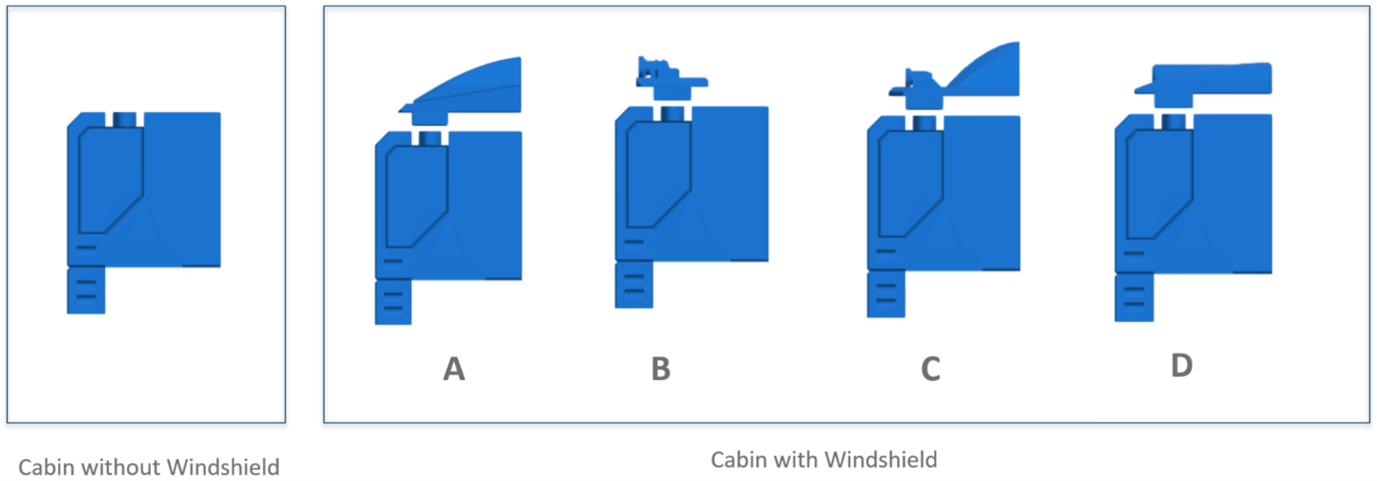}
    \caption{Illustration of the Cabin with  windshield (4 different types) and without windshield use-case.}
    \label{fig:cabin and windshield}
\end{figure}

The use-case of quality inspection USB sticks in manufacturing \cite{Hegiste.2022} was extended via annotating the previous dataset in YOLO format \cite{Redmon.2016}, as required by the YOLOv5 algorithm.
To further extend the usability of this algorithm, we introduce a new use-case where two companies/clients produce cabins and windshields as mentioned in Figure \ref{fig:cabin and windshield}. 
The quality inspection use-case here is to detect whether the cabin is with or without a windshield.
The dataset and further details will be explained in subsection \ref{datasets}.

\subsection{Datasets}\label{datasets}
\begin{figure}[h]
    \centering
    \includegraphics[width= 0.48\textwidth]{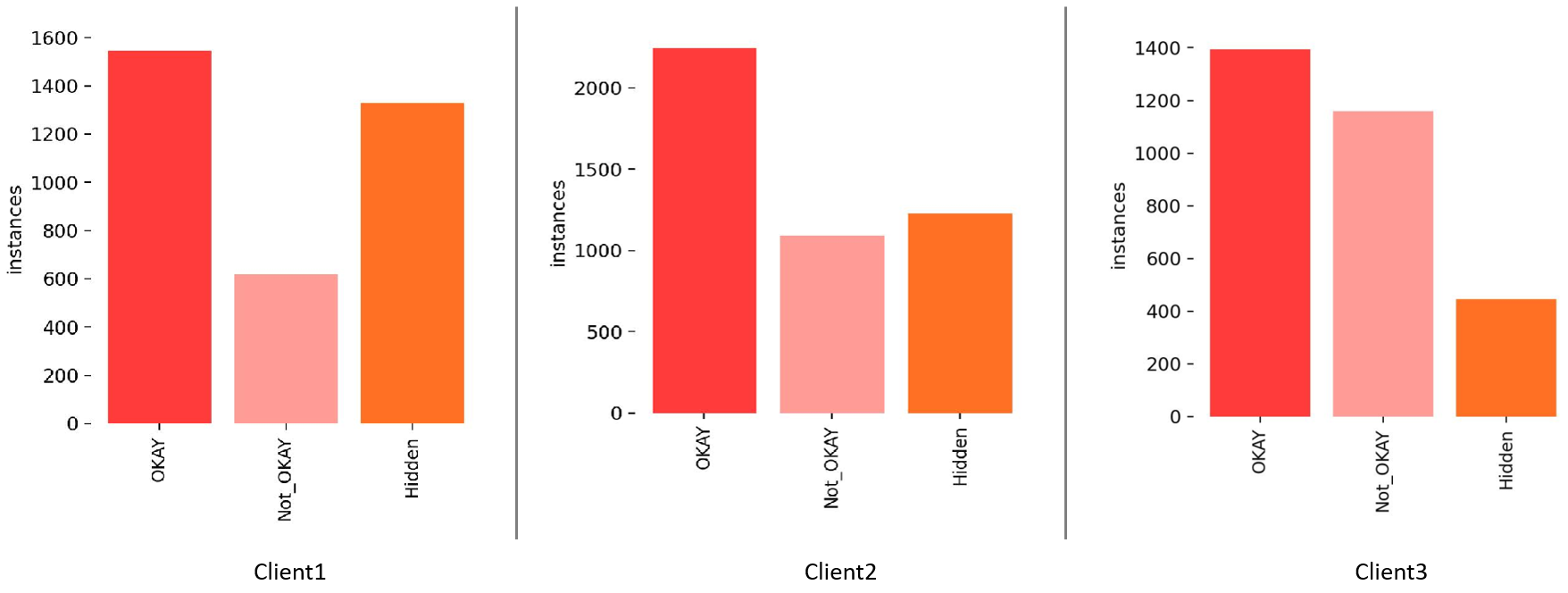}
    \caption{Training dataset distribution and label instances of client1 (Huawei on left), client2 (SF blue in middle) and client3 (SF red on right)}
    \label{fig:USB data instances}
\end{figure}

\begin{figure}[h]
    \centering
    \includegraphics[width= 0.48\textwidth]{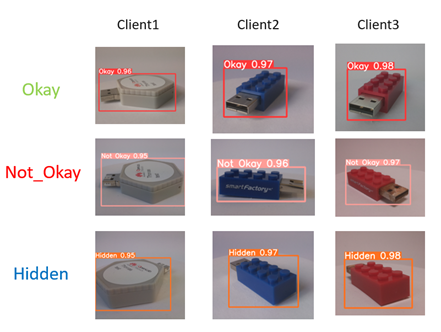}
    \caption{Small subset example of USB quality inspection dataset, client1 (Huawei on left), client2 (SF blue in middle) and client3 (SF red on right)}
    \label{fig:USB dataset anno}
\end{figure}

In this subsection, we will explain both USB and cabin quality inspection use-case.
Starting with USB quality inspection, we have three clients and each of them has three classes consisting of 'Okay', 'Not\_Okay' and 'Hidden' (similar to \cite{Hegiste.2022}). The dataset is non-independent and identically distributed (non-IID), and the dataset distribution along with their respective classes can be seen in Figure \ref{fig:USB data instances}. 
Client1 has the Huawei USB sticks, client2 has a blue brick style USB stick and client3 has a red brick style USB stick. Each client consists of three classes, where the 'Not\_OKAY' class of each client has a different error type.
As shown in Figure \ref{fig:USB dataset anno}\footnote{The object in the images are zoomed a little for clear view}, client1's USB error is small sticker marks on the USB port which illustrates scratch, client2's USB error is the USB port being damaged and client3's error is a rusted USB port. 
Similar to the use-case mentioned in paper \cite{Hegiste.2022}, where we showcased successful federated image classification for custom quality inspection, in this paper the purpose is to see that if federated OD algorithm could achieve the same results where the global federated model is able to learn all different types of USB errors and more importantly draw a perfect bounding box over the object. The detailed part regarding FedOD algorithm is mentioned in subsection \ref{methodology}.

\begin{figure}[h]
    \centering
    \includegraphics[width= 0.45\textwidth]{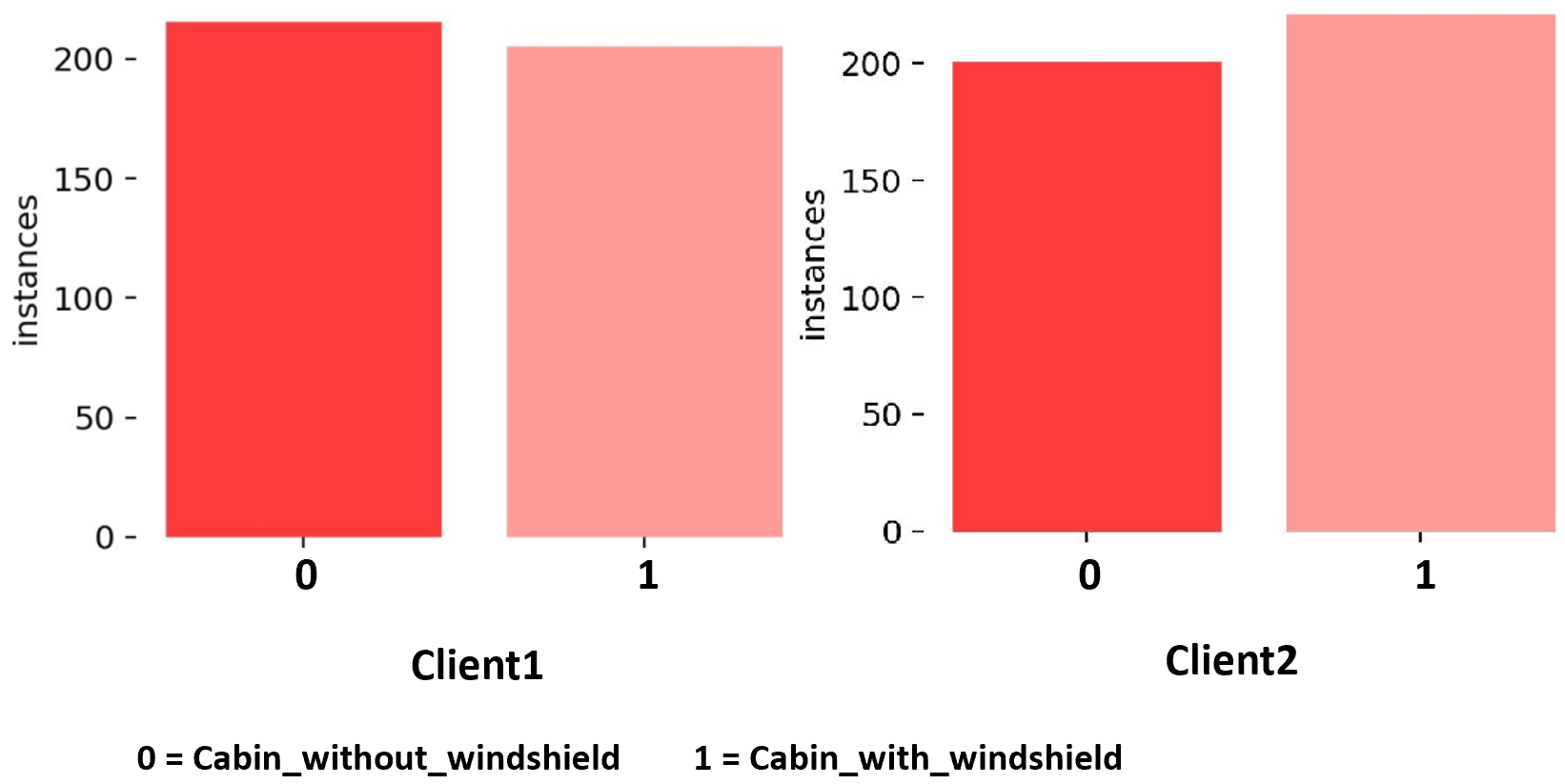}
    \caption{Training dataset distribution and label instances of client1 (Blue cabin on left), client2 (Red cabin on right)}
    \label{fig:Cabin data instances}
\end{figure}

\begin{figure}[h]
    \centering
    \includegraphics[width= 0.49\textwidth]{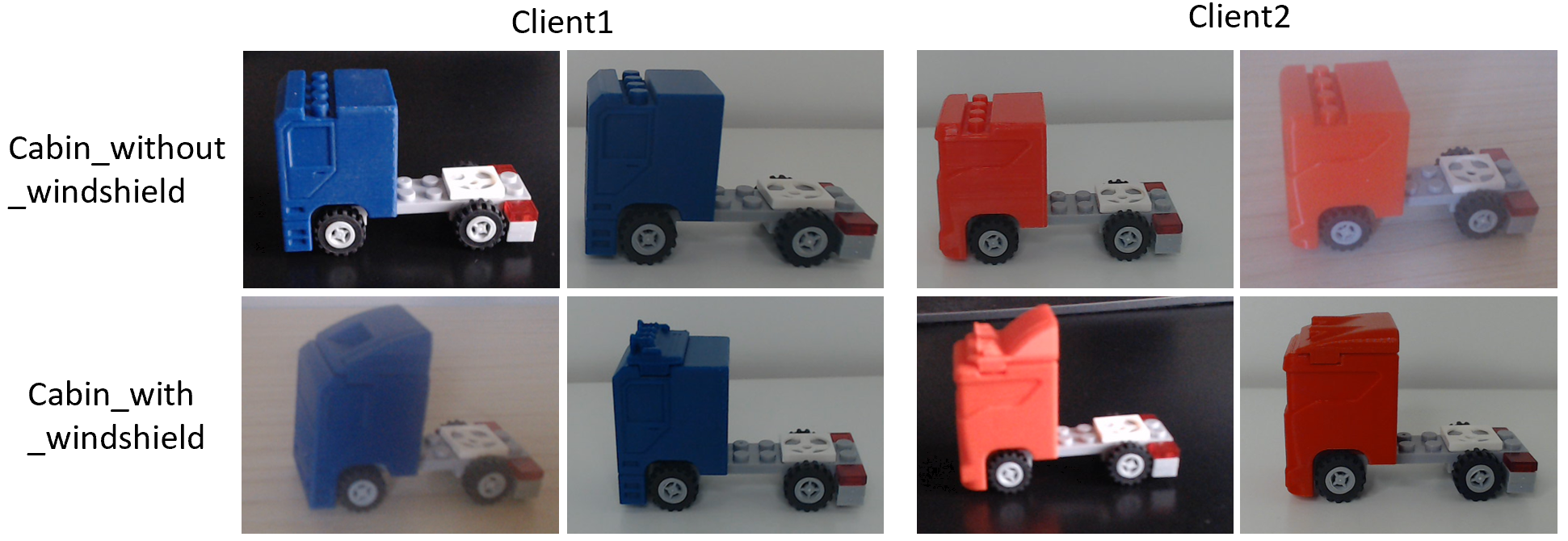}
    \caption{Small subset of the cabin quality inspection dataset, client1 (Blue cabin on left), client2 (Red cabin on right)}
    \label{fig:cabin_data}
\end{figure}

For the second use-case, there are two companies/clients which manufacture cabins with and without windshield. The main application is to create an object detector model to classify and detect the object correctly in a given video or image.
We refer to the companies as client1 and client2, where client1 only manufactures blue colored cabins and windshield of type A and B and client2 manufactures red colored cabins (a little different design than the blue colored cabins) and windshield of type C and D (refer to Figure \ref{fig:cabin_data}).
The classes instance for each clients' training dataset can be seen in Figure \ref{fig:Cabin data instances}. Each client had a total of approximately 600 images, out of which 15\% each is used for validation and testing.
The dataset was created with the cabin installed in a chassis (but in the annotation the bounding box was only drawn over the cabin) and with 3 different backgrounds. Various parameters such as different lighting conditions, shadows, blurry images, etc. were also introduced via creation of the custom dataset.
The federated approach of these use-cases is explained further in subsection \ref{imple and exp}

\subsection{Implementation} \label{imple and exp}
\begin{figure}[h]
    \centering
    \includegraphics[width= 0.45\textwidth]{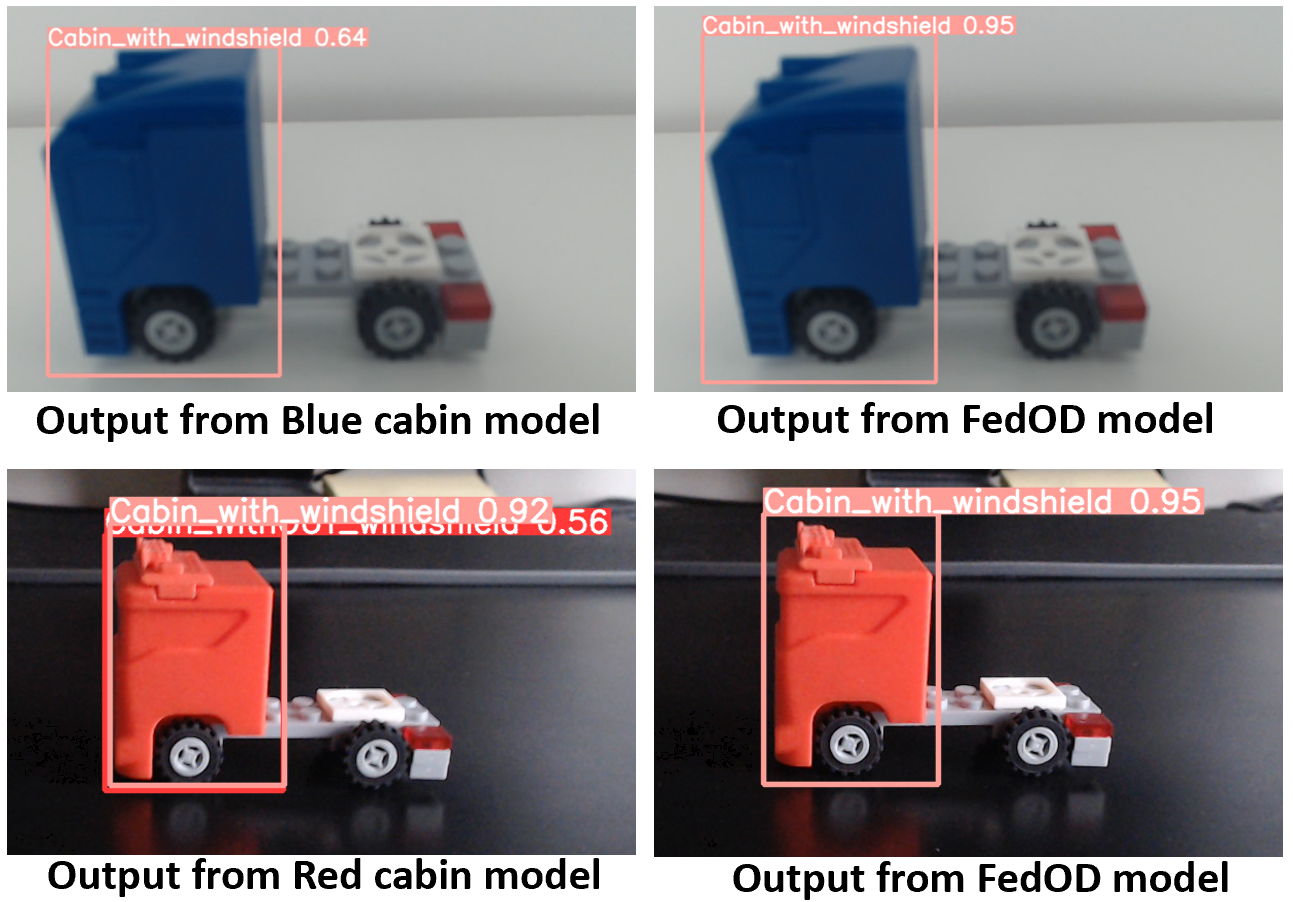}
    \caption{Output of models trained on local dataset, client1 (left upper) and client2 (left lower) and global federated YOLOv5 model (right column) on unseen windshield type dataset.}
    \label{fig:clients_vs_global}
\end{figure}

As previously mentioned, there are two clients involved in the manufacturing of cabins and windshields with different designs and types. For their local quality inspection models, they utilize their respective local datasets (see Figure \ref{fig:cabin_data}) to train YOLOv5 models that can detect cabins with or without windshields in a given frame.
Both clients' models achieve more than 95\% accuracy on their own local test datasets. Client1's model is evaluated on blue cabins without windshields and with windshields of type A and B, while client2's model is tested on cabins without windshields and with windshields of type C and D.
In the future, both clients plan to manufacture the remaining two windshields, each with its own color type. This means that client 1 will produce cabins with windshield type C and D, while client 2 will manufacture cabins with windshield type A and B, in addition to their existing production.
The clients' locally trained models, based on their old datasets, were tested with the new cabin-windshield combinations and can be referred in Figure \ref{fig:clients_vs_global}. However, the results show that although the local models correctly classify the images as either 'Cabin\_without\_windshield' or 'Cabin\_with\_windshield,' the bounding boxes generated are imprecise and sometimes crop part of the windshield. In some cases, false positives are detected, and labels are assigned with lower confidence scores, as depicted from the bottom-left of Figure \ref{fig:clients_vs_global}.
While the clients could potentially share their datasets to train a centralized YOLOv5 model for quality inspection, personal or competitive reasons prevent them from sharing their raw local image data. Consequently, both clients would need to create new additional data for the new combinations, annotate the dataset, and re-train the entire model to classify the new 'Cabin\_with\_windshield' images.
However, the process of manually creating and annotating the dataset for each client is tedious. This is where FL plays a crucial role, enabling the development of a final global model that can accurately detect objects from both clients without the need to share their raw image data.
\begin{figure}[h]
    \centering
    \includegraphics[width= 0.4\textwidth]{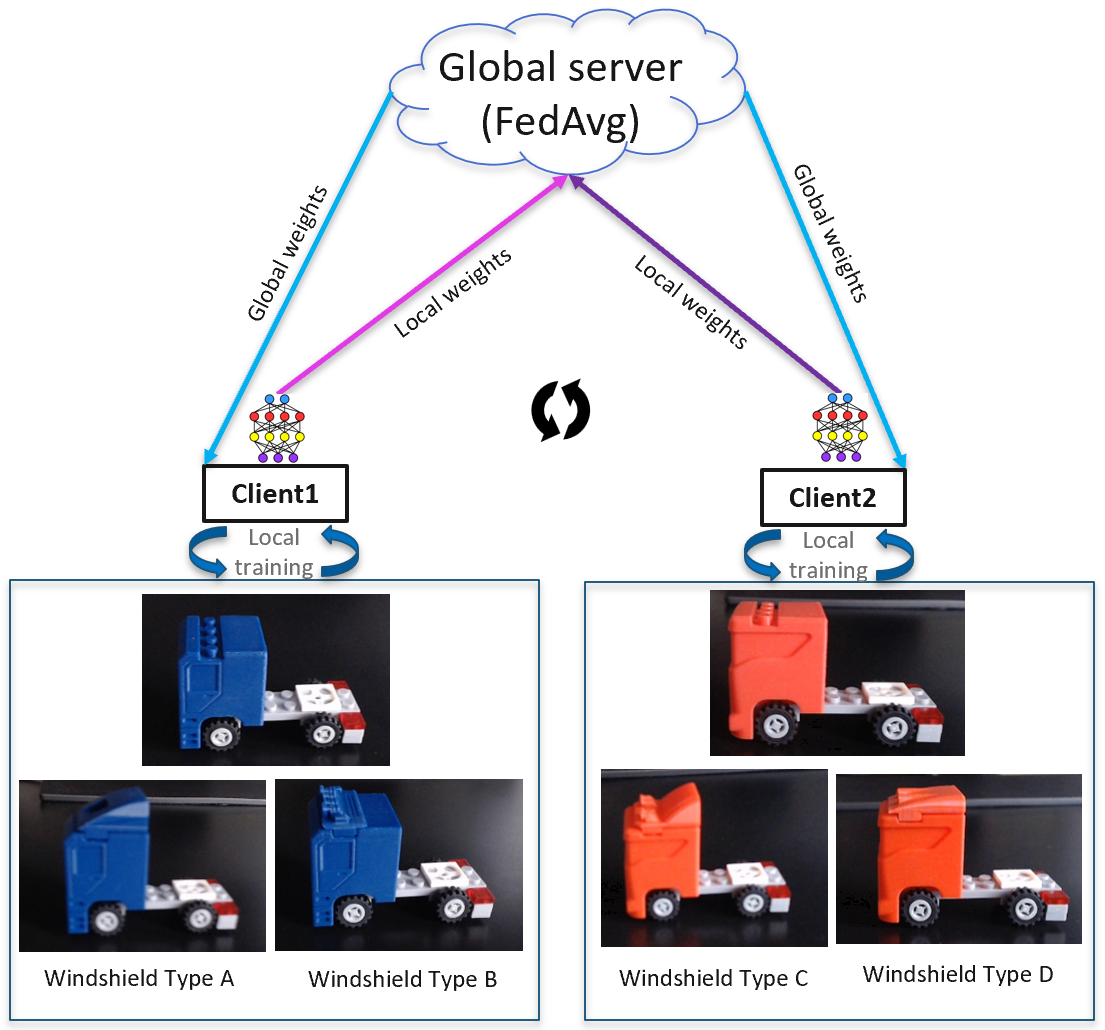}
    \caption{Illustration of federated learning between cabin clients}
    \label{fig:fedod clients}
\end{figure}

The FL algorithm employed in our work is FedAvg [6].A neutral server is utilized to perform federated averaging on the model weights from all clients. Certain assumptions are made, such as including the active participation of all clients in each communication round and also their trustworthiness in sharing their model weights.
Before commencing the training process, all clients agree on a standardized label nomenclature, the YOLO model architecture, and other hyperparameters such as local epochs, optimizer, and batch size. They then initiate their respective local training procedures. Figure \ref{fig:fedod clients} shows a pictorial view of this federated learning process.
Once the local training is completed, the model weights achieved by each client are sent to the neutral server. After receiving the model weights from all clients, the server performs federated averaging and sends the updated global weights back to each client. This process is referred to as one communication round (CR).
Through multiple CRs, the global model gradually improves and demonstrates enhanced performance on both clients' test datasets. In this particular use-case, each client runs the global model received from the server on its local test dataset, providing feedback on the accuracy of the global model and transmitting new local weights accordingly.
The average accuracy across all clients' test datasets serves as the stopping parameter. Once the server receives the accuracy of the previous global model on all the clients' local test datasets and calculates the average accuracy to be greater than 96\% (as this was the accuracy of the normally trained YOLOv5 model), the previous global weights are transmitted to the clients as the final global federated model.
In this particular use-case, the global model was achieved after 10 CRs with 15 local epochs. The output of this model on a similar test dataset demonstrated very high accuracy, with highly precise bounding boxes around the objects. Figure \ref{fig:clients_vs_global} showcases the output of the global federated model on similar images. It can be observed that the confidence scores of the predictions are very high, and the bounding boxes accurately encompass the windshields. There are no false positive detections, and the model output exhibits robustness even for blurry images.
A similar setting was used for the USB quality inspection dataset (Figure \ref{fig:USB dataset anno}), where the final global model was able to correctly classify various errors and precisely draw bounding boxes around specific USB sticks. The global federated USB stick model was even able to detect a combination never seen in the dataset, such as client 1's sticker error when applied to client 2's USB sticks.
This aligns with the results presented in the paper \cite{Hegiste.2022}, which focused on a federated image classification setting.

\subsection{Experimentation} \label{exp}

In this paper, the focus is primarily on the cabin quality inspection use-case, and we conducted experiments based on this scenario. While a quick overview of the results for the USB federated OD model can be seen in Figure \ref{fig:4model_usb}, we will specifically concentrate on evaluating the performance of different models in the cabin quality inspection domain. Throughout the experiments, we will refer to client1's locally trained model as the "Blue cabin model," client2's model as the "Red cabin model," and the global federated model as "FedOD."

The following experiments were conducted:
\begin{enumerate}
    \item Testing the Blue cabin model, Red cabin model, and FedOD model on the new cabin-windshield combinations as test dataset.
    \item Evaluating all three models for live object detection with various combinations of cabins and windshields.
    \item Testing all three models on images obtained from the quality inspection module of the manufacturing process.
\end{enumerate}
In the first experiment, the test dataset consisted of blue cabins with windshields of type C and D, as well as red cabins with windshields of type A and B. These combinations were not present in the local training datasets of either client. The goal is to assess the performance of these models on these previously unseen combinations.
For the second experiment, all three models (Blue cabin, Red cabin, and FedOD) were validated simultaneously to detect objects in live frames with different combinations of cabins and windshields. This experiment aims to compare the outputs of the models under various scenarios, including frames with multiple cabins.
Lastly, in the third experiment, images from the quality inspection module situated at SmartFactory-Kaiserslautern (SF-KL) will be used as input for the models. It's important to note that the background and lighting conditions of these images differ significantly from those in the training dataset, adding a new level of complexity to the evaluation process.

\subsection{Integration with the Industry 4.0 Shared Production architecture}
\begin{figure}[h]
    \centering
    \includegraphics[width= 0.5\textwidth]{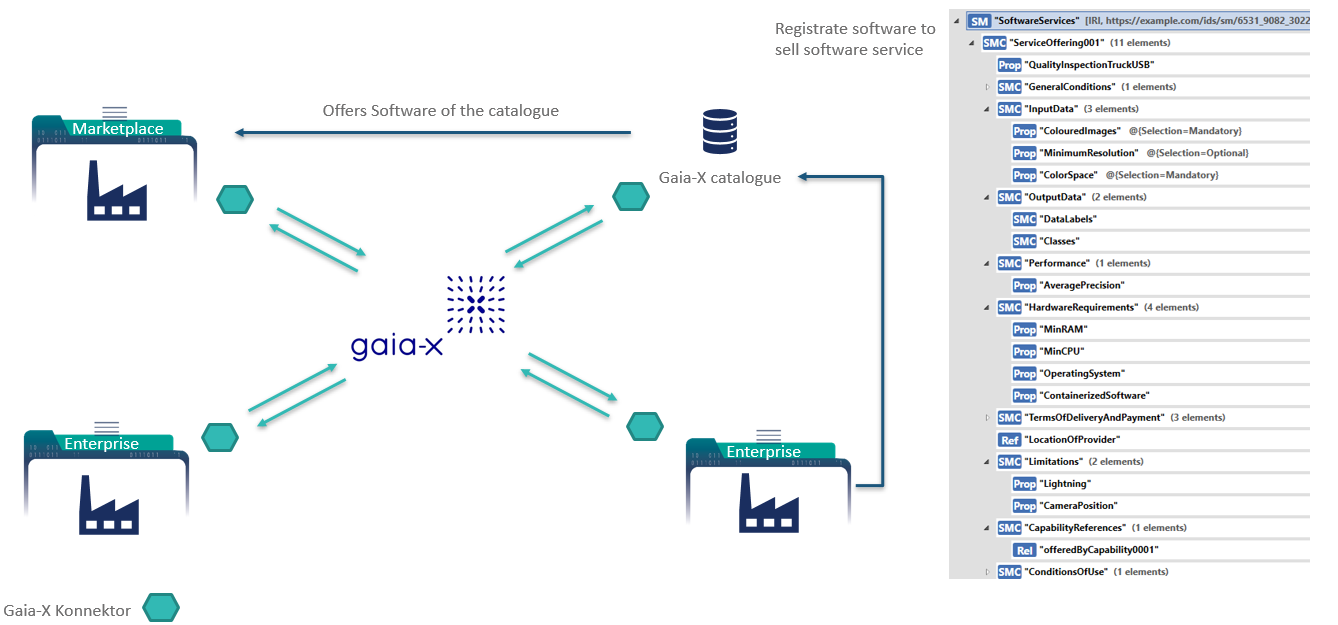}
    \caption{Industry 4.0 data space for software services \cite{Simon.2023}}
    \label{fig:AAS_GAIAX}
\end{figure}

In this subsection, we explore and give one possible solution how we can offer our quality inspection AI-software service to different companies over Gaia-X platform. Therefore, we orientated on the offering of production services in a Gaia-X data spaces published in \cite{Jungbluth.2023}, \cite{Simon.2023} and \cite{Volkmann.2023}. In this case, a potential user of the service can download and use the service with own dataset in own production line without any significant efforts. To offer the quality inspection service in Gaia-X we have described the quality inspection service (its features, characteristics, properties, and capabilities) in a submodel of AAS. Currently, there is no standard submodel template available to describe AI-service capabilities \cite{IDTA.2023}, so to create the submodel, we used Google model cards \cite{Google.2023} as an example of how AI software services can be described in a common way. 
With the help of Gaia-X connector, we can connect to the related data space and offer our AAS-based description of the software service. That means it can be found in the Gaia-X service catalog and the service provider can offer the software service via marketplace (Figure \ref{fig:AAS_GAIAX}). 

If the customer connects to the data space, he can browse available software services on the marketplace, select one, which matches his requirements, download it (as a Docker container) and use in his production line. All this operation is performed with the help of service generalized description available via AAS. The customer can download the service or, if the service is already running on the customer side, just update the model weights from the global federated model. 
Because the quality inspection service based on FL algorithm, the customer has also an opportunity to contribute to the service by improving the model quality via an additional round of training on his local dataset. But the main challenge here is to make sure that all customers have the similar data classes and use case. Each FL model is trained for a specific use case, and precise description of possible use cases is necessary to offer a product on a marketplace. Otherwise, the model quality can be corrupted by ingesting the wrong local datasets. Although some precautions, such as attempting an automatic assignment of class labels from different clients, can be taken \cite{Legler.2023}; however, a trustworthy environment must be in place to ensure cooperation.

\section{Results and Discussion} \label{results_diss}
\begin{figure}[h]
    \centering
    \includegraphics[width= 0.48\textwidth]{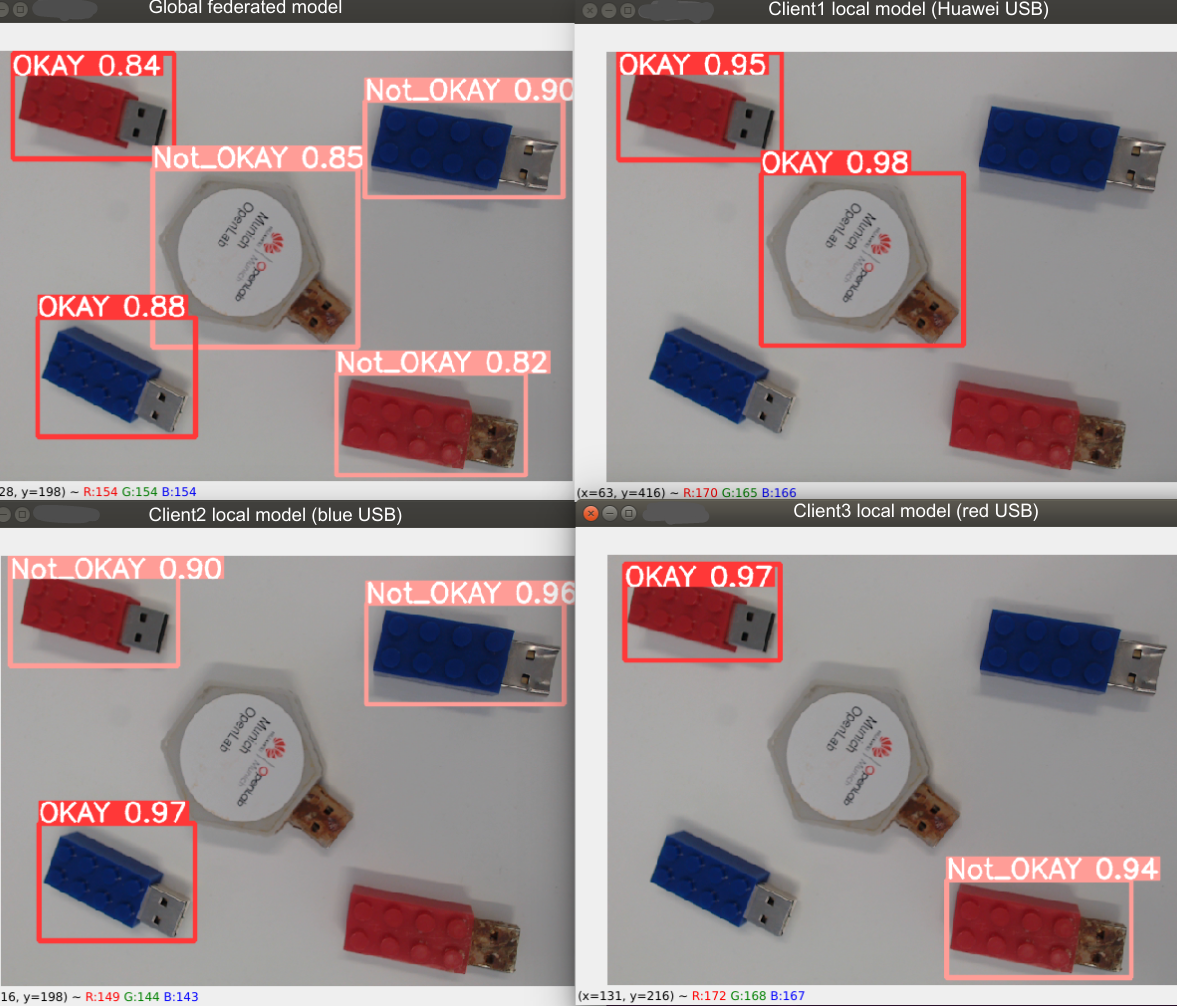}
    \caption{Live comparison of federated global model with models trained on their local respective dataset}
    \label{fig:4model_usb}
\end{figure}

In this section, we present the results of federated USB quality inspection and results of our experiments conducted for the cabin quality inspection use-case, which are described in detail in subsection \ref{exp}. We focus on three key experiments to compare the performance of three different models.
Figure \ref{fig:4model_usb}, shows 4 windows with output from each model, namely the global federated model which was achieved by 15 local epochs for 5 CR, followed by client1, client2 and client3 models which were trained for 150 epochs based on their local dataset. We can see that the global FedOD model is not only able to predict all clients' error types, but also detect error of client3 (rust) on client1's USB stick.

Moving on to the experiments, in the first experiment, we evaluated these models on a test dataset that included both blue and red cabins with different windshield combinations. It was observed that the Blue cabin model was unable to accurately classify or detect the red-colored cabins and windshields. Similarly, the Red cabin model struggled to detect the blue-colored cabins and windshields, producing incorrect bounding boxes for those images. In contrast, the FedOD model demonstrated superior performance by successfully detecting all different cabin and windshield combinations, generating highly precise bounding boxes for most test images. Detailed results for this experiment can be found in Table \ref{table1}.
Table \ref{table1} presents the average precision (AP) values for different IoU thresholds ranging from 0.50 to 0.95, as well as the mean average precision (mAP) at an IoU threshold of 0.5. The FedOD model achieved an AP\@[.50:.05:.95] of 0.93 and an mAP of 1.0, showcasing its robust performance across a wide range of IoU thresholds.
To further investigate the accuracy of the Blue cabin and Red cabin models, we conducted a second experiment where each model was tested specifically on its corresponding cabin and windshield color combination. Additionally, we tested the FedOD model on both combinations to enable a direct comparison. The results of this experiment are presented in Table \ref{table2}.
The mAP and AP\@[.50:.05:.95] values in Table \ref{table2} indicate that the FedOD model outperformed the local models in terms of precision. It consistently achieved higher mAP and AP scores, suggesting that the global FedOD model excels in predicting precise bounding boxes, even when confronted with unseen combination types.
These results highlight the effectiveness of the FedOD model in the cabin quality inspection use-case, demonstrating its ability to accurately detect different cabin and windshield combinations which were not even the part of its local training datasets. The superior performance of the FedOD model provides strong evidence for the advantages of FL in collaborative object detection scenarios.

\begin{table*}
\centering
\caption{mAP metrics comparison of Blue cabin, Red cabin and FedOD model on an unseen Test dataset (AP= Average precision, APm = AP of medium-sized objects, APl= AP of large-sized objects, AR = Average Recall)}
\begin{tabular}{|c|l|l|l|c|l|l|l|l|}
\hline
Model & \multicolumn{1}{c|}{Training Dataset} & Test Dataset & mAP & \multicolumn{1}{l|}{AP@{[}.50:.05:.95{]}} & APm & APl & ARm & ARl \\ \hline
\begin{tabular}[c]{@{}c@{}}Client1\\ (Blue cabin)\end{tabular} & \begin{tabular}[c]{@{}l@{}}Blue colored cabins without \\ windshield and cabin with\\ windshields of type A and B\end{tabular} & \multirow{3}{*}{\begin{tabular}[c]{@{}l@{}}Blue colored cabins with windshield\\ type C and D and Red colored cabins\\ with windshield type A and B. (Does \\ not belong to any training dataset)\end{tabular}} & 0.42 & 0.35 & \_\_ & 0.35 & \_\_ & 0.36 \\ \cline{1-2} \cline{4-9} 
\begin{tabular}[c]{@{}c@{}}Client2\\ (Red cabin)\end{tabular} & \begin{tabular}[c]{@{}l@{}}Red colored cabins without\\ windshield and cabin with\\ windshields of type C and D\end{tabular} &  & 0.49 & 0.42 & \_\_ & 0.42 & \_\_ & 0.43 \\ \cline{1-2} \cline{4-9} 
\begin{tabular}[c]{@{}c@{}}\textbf{Global Federated}\\ \textbf{model}\end{tabular} & \multicolumn{1}{c|}{\_\_} &  & \textbf{1.0} & \textbf{0.93} & \_\_ & \textbf{0.93} & \_\_ & \textbf{0.96} \\ \hline
\end{tabular}
\label{table1}
\end{table*}

\begin{table*}
\centering
\caption{The centralized trained YOLOv5 models (client1 and client2) trained on their local dataset and FedOD model are tested with Windshield combination not present in their training dataset}
\begin{tabular}{|c|c|c|c|c|c|c|c|c|}
\hline
Model & Training Dataset & Test Dataset & mAP & AP@{[}.50:.05:.95{]} & APm & APl & ARm & ARl \\ \hline
\begin{tabular}[c]{@{}c@{}}Client1\\ (Blue cabin)\end{tabular} & \begin{tabular}[c]{@{}c@{}}Blue cabins and\\ windshield of\\ type A and B\end{tabular} & \begin{tabular}[c]{@{}c@{}}Blue cabins with \\ windshield type\\ C and D\end{tabular} & 0.83 & 0.70 & \_\_ & 0.70 & \_\_ & 0.73 \\ \hline
\textbf{\begin{tabular}[c]{@{}c@{}}Global federated\\ model\end{tabular}} & \textbf{\_\_\_} & \begin{tabular}[c]{@{}c@{}}Blue cabins with \\ windshield type\\ C and D\end{tabular} & \textbf{1.0} & \textbf{0.96} & \textbf{\_\_} & \textbf{0.96} & \textbf{\_\_} & \textbf{0.98} \\ \hline
\begin{tabular}[c]{@{}c@{}}Client2\\ (Red cabin)\end{tabular} & \begin{tabular}[c]{@{}c@{}}Red cabins and\\ windshield of \\ type C and D\end{tabular} & \begin{tabular}[c]{@{}c@{}}Red cabins with\\ windshield type\\ A and B\end{tabular} & 0.97 & 0.83 & \_\_ & 0.83 & \_\_ & 0.86 \\ \hline
\textbf{\begin{tabular}[c]{@{}c@{}}Global federated\\ model\end{tabular}} & \_\_\_ & \begin{tabular}[c]{@{}c@{}}Red cabins with \\ windshield type\\ A and B\end{tabular} & \textbf{1.0} & \textbf{0.91} & \textbf{\_\_} & \textbf{0.91} & \textbf{\_\_} & \textbf{0.93} \\ \hline
\end{tabular}

\label{table2}
\end{table*}

\begin{figure}[h]
    \centering
    \includegraphics[width= 0.48\textwidth]{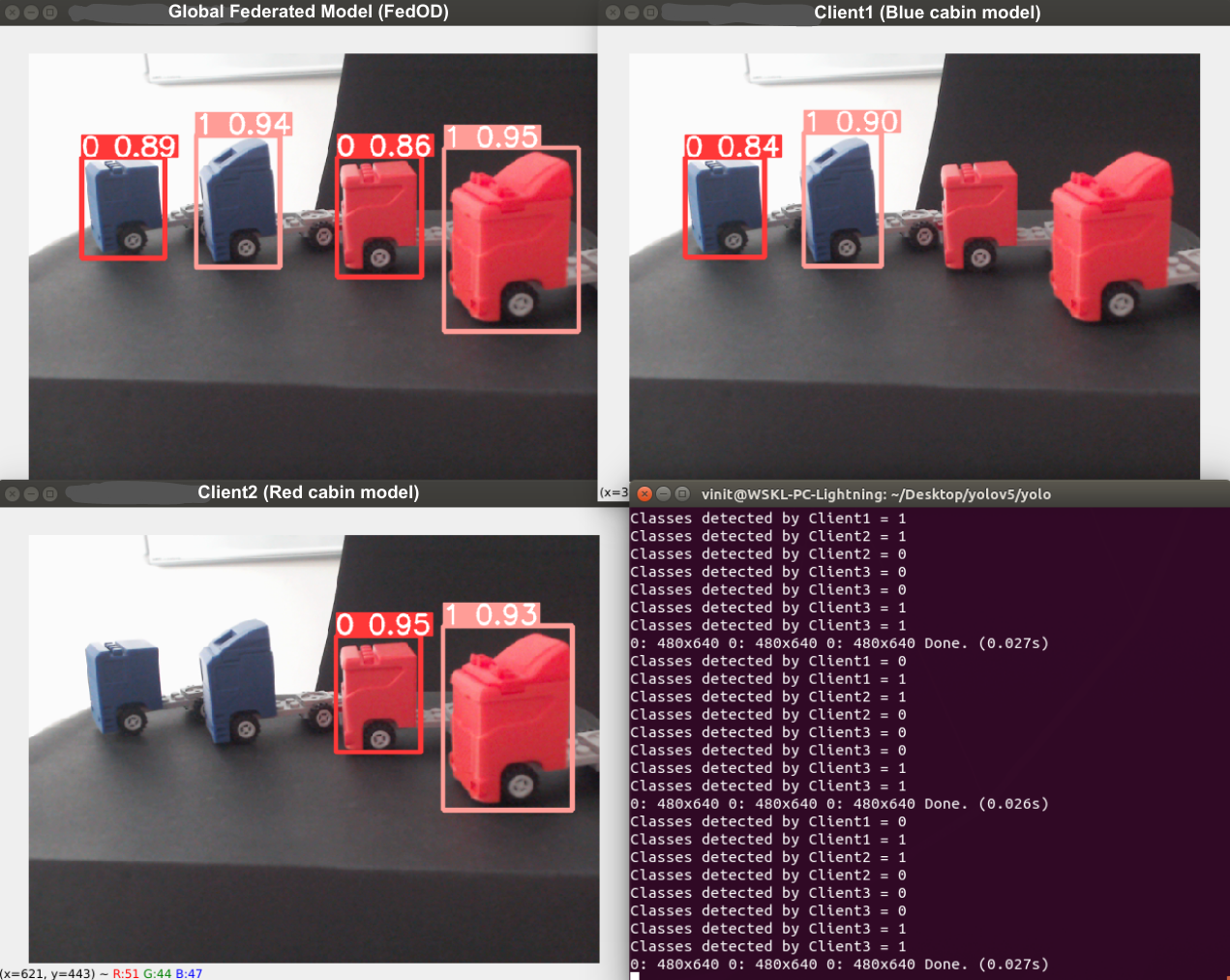}
    \caption{Comparison of federated global model vs. models trained with their local data using live object detection (0: Cabin\_without\_windshield, 1: Cabin\_with\_windshield)}
    \label{fig:Comparison1}
\end{figure}

\begin{figure}
    \centering
    \includegraphics[width= 0.48\textwidth]{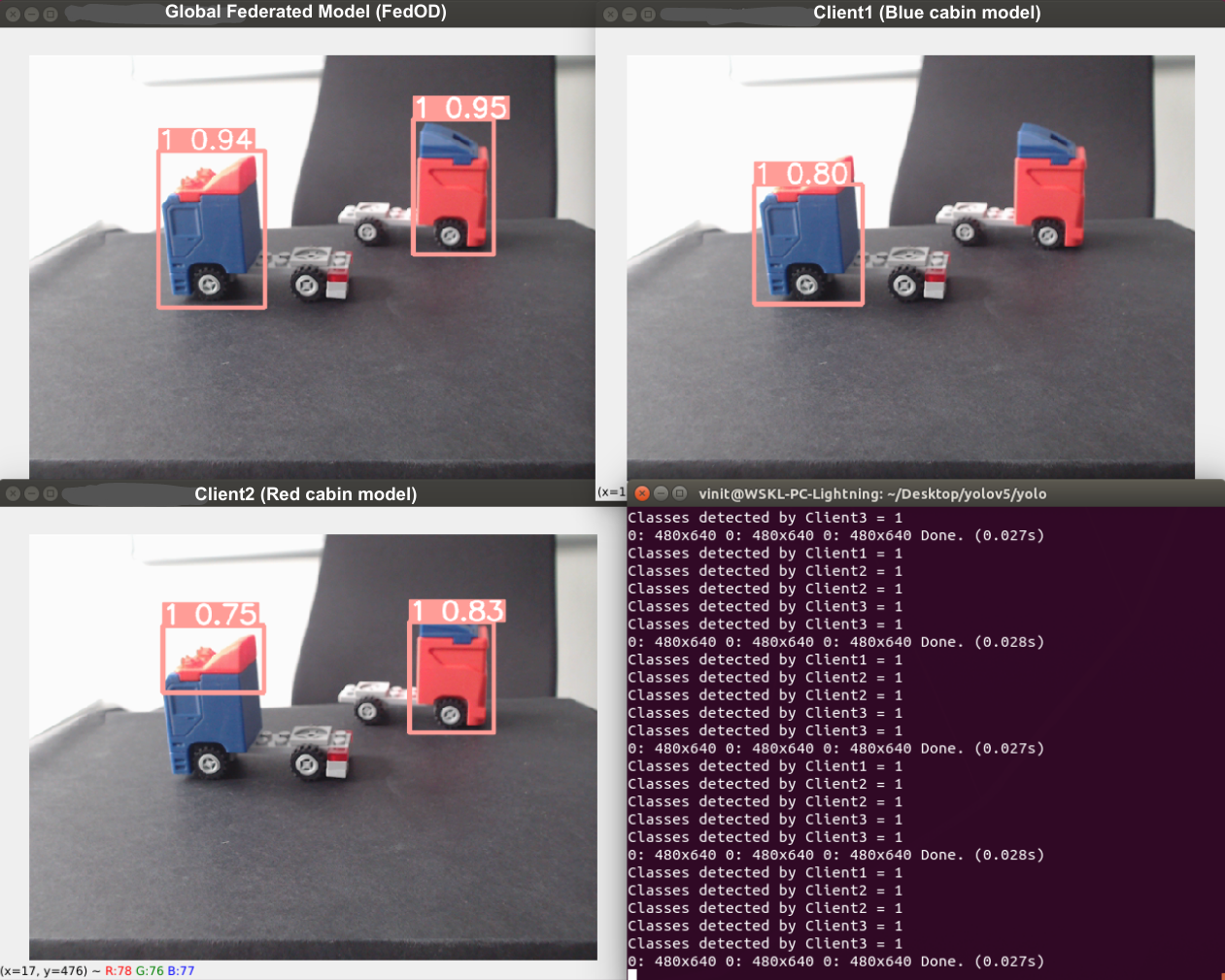}
    \caption{Comparison of federated global model vs. models trained with their local data using live object detection}
    \label{fig:Comparison2}
\end{figure}

In the second experiment, we developed a custom code to simultaneously run all three models (Blue, Red and FedOD cabin model) in parallel on live video streams. This setup allowed us to directly compare the output of each model and observe any discernible differences.
Multiple cabin combinations were tested within a single frame, as depicted in Figure \ref{fig:Comparison1} and Figure \ref{fig:Comparison2}. Each figure consists of three windows. The top-left window displays the output of the global FedOD model. 
The top-right window showcases the Blue cabin model, which was trained using the local dataset of client1, and the bottom-left window presents the output from the Red cabin model, trained on the local dataset of client2.
Figure \ref{fig:Comparison1} and Figure \ref{fig:Comparison2}, follow the same pattern of windows and the labels output are changed to 0 and 1 for representing 'Cabin\_without\_windshield' and 'Cabin\_with\_windshield' respectively, to provide a clear visual comparison of the models' outputs across different bounding boxes.

In Figure \ref{fig:Comparison1}, the frame contains 4 cabins with the combination available in the training dataset. The FedOD model demonstrates superior performance by accurately detecting all 4 cabins with high confidence for each client's combination, while it is evident that the Blue cabin model fails to detect the red cabin and vice versa for the Red cabin model. Both models correctly identify only their own design type, which proves the low performance of these models in Table \ref{table1}. 
Moving on to Figure \ref{fig:Comparison2}, a red-designed cabin from client2 with windshield type A from client1 was tested. The results are remarkable: the Red cabin model successfully classifies the object with a high confidence score, but the bounding box drawn is imprecise and crops a portion of the windshield. On the other hand, the Blue cabin model fails to detect this specific object entirely. 
Figure \ref{fig:Comparison2} also presents a reverse scenario of the previous test case, where a blue cabin with windshield type C from client2 was tested. The Blue cabin model correctly classifies the object, but similar to the previous scenario, it struggles to generate an accurate bounding box. The output of the Red cabin model is intriguing: since windshield type C was part of its training dataset, it appears to classify the object as 'Cabin\_with\_windshield', although with an incorrect bounding box. In contrast, the FedOD model not only accurately classifies both the objects, but also draws a precise bounding box over these unseen combination types.

\begin{figure}[h]
    \centering
    \includegraphics[width= 0.48\textwidth]{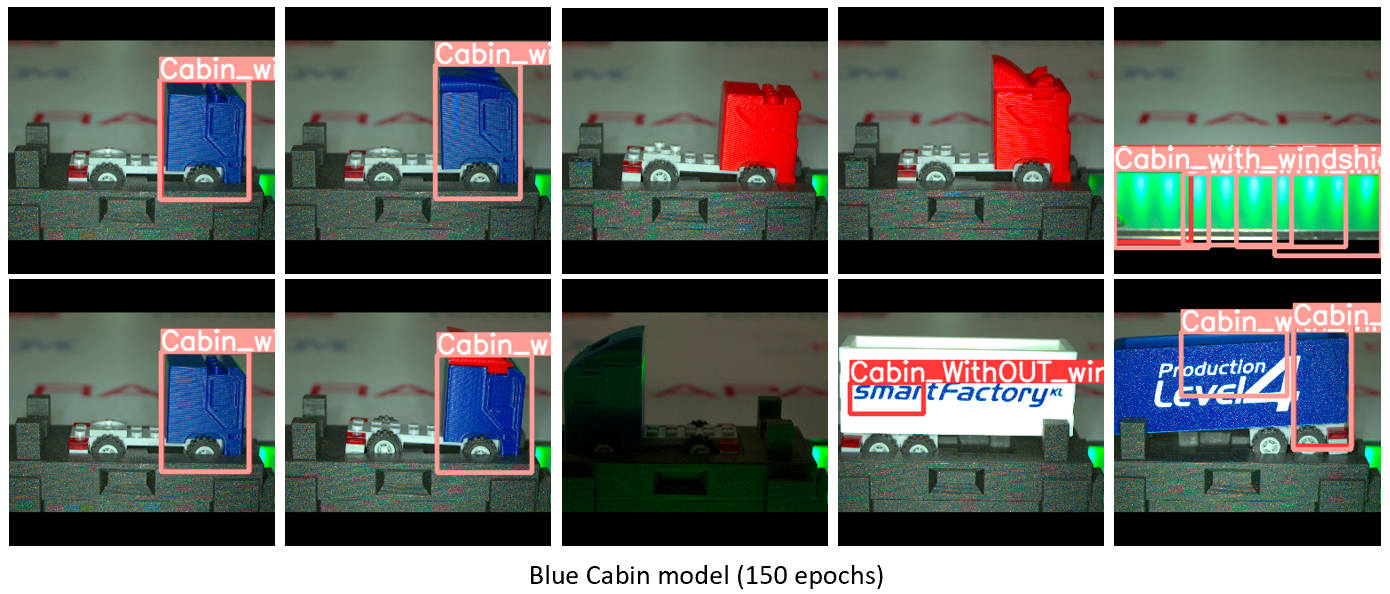}
    \caption{Output of model trained with only blue cabin dataset (Blue cabin model) on images from quality inspection of demonstrator}
    \label{fig:Demonstator_blue}
\end{figure}

\begin{figure}[h]
    \centering
    \includegraphics[width= 0.48\textwidth]{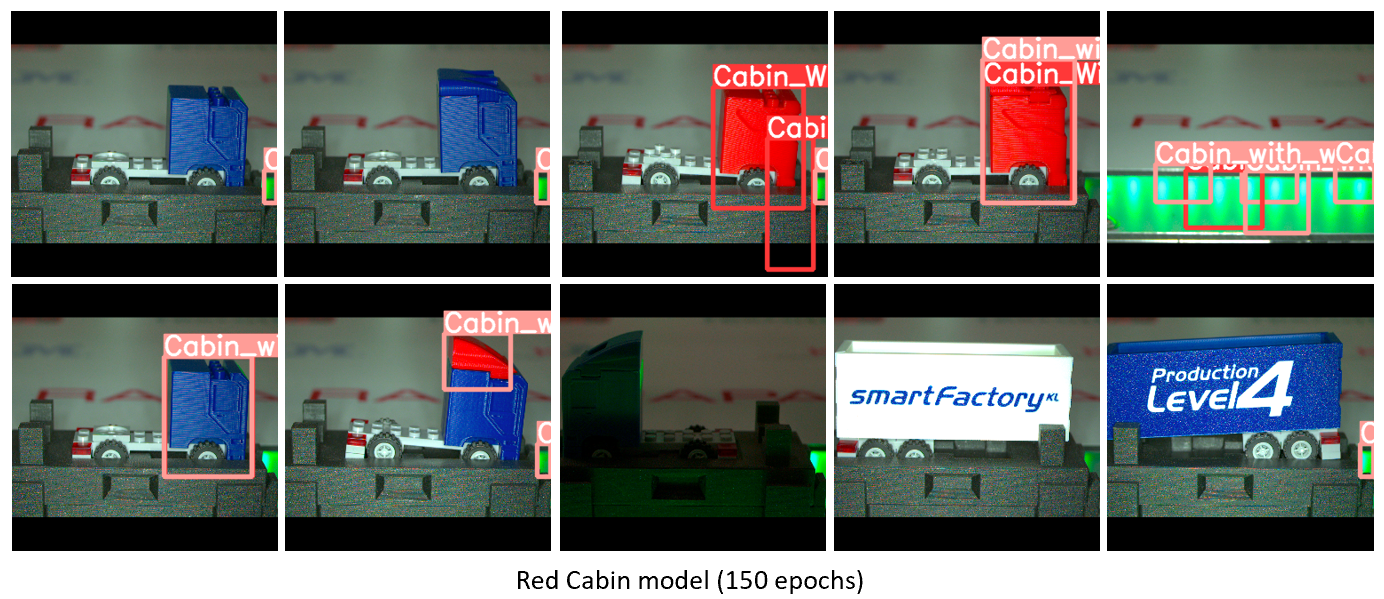}
    \caption{Output of model trained with only red cabin dataset (Red cabin model) on images from quality inspection of demonstrator}
    \label{fig:Demonstator_red}
\end{figure}

\begin{figure}[h]
    \centering
    \includegraphics[width= 0.48\textwidth]{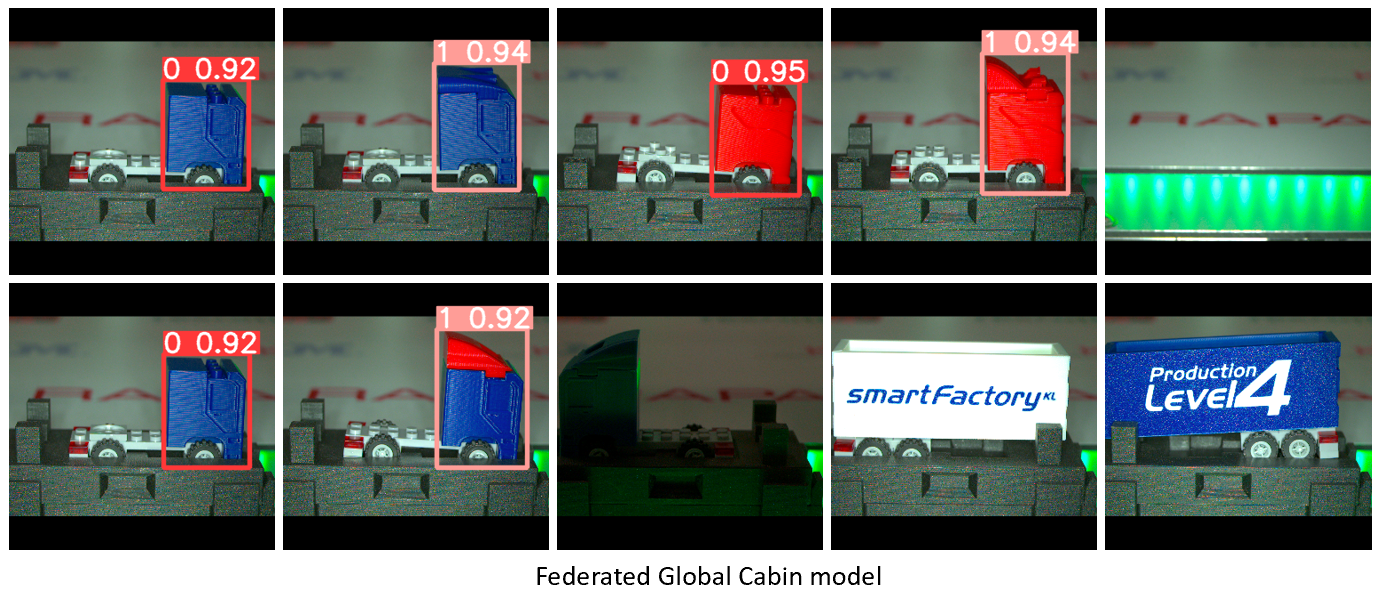}
    \caption{Output of global federated model (FedOD model) on images from quality inspection of demonstrator (0: Cabin\_without\_windshield, 1: Cabin\_with\_windshield)}
    \label{fig:Demonstator_fed}
\end{figure}

The third experiment involved testing the same models on images obtained from the demonstrator located at SF-KL in the quality inspection module, as described in subsection \ref{exp}. Figure \ref{fig:Demonstator_blue}, \ref{fig:Demonstator_red}, and \ref{fig:Demonstator_fed} showcase the results obtained from these tests.
It is worth noting that the images captured in this setting have significantly different background and lighting conditions compared to the images used in the training dataset. The consistent use of the same set of images in all three figures allows for a direct comparison of the output generated by each model.
In Figure \ref{fig:Demonstator_blue}, the output of the Blue cabin model reveals its inability to correctly classify instances of 'Cabin\_without\_windshield'. Additionally, the model predicts numerous false positives on the trailer objects and even in frames where no objects are present.
Similarly, Figure \ref{fig:Demonstator_red} illustrates the performance of the Red cabin model, which also struggles to accurately detect objects in the images from the quality inspection module. The model exhibits misclassifications and false positives, particularly on the trailers and frames without any objects.
The results obtained from our global FedOD model on the same test images are truly remarkable. The enhanced algorithm, which combines the power of FL and OD, has demonstrated substantial improvements in accuracy and precision compared to the previous individual client models. Figure \ref{fig:Demonstator_fed}, shows the output of FedOD model on the test images. 
The figure reveals the FedOD model's exceptional performance in predicting bounding boxes with remarkable precision and confidence scores. 
Notably, the model exhibits no false positives when detecting different types of trailers or when confronted with frames containing just the background and no objects. These findings provide compelling evidence of the versatility and generalizability of our FedOD model, particularly in detecting identical objects in diverse and previously unseen environments, as well as various object combinations which are unseen by the training model.


In our future work we will focus on implementation of the standard AAS sub-model for AI-service capabilities description that should be released soon according to Industrial Digital Twin Association\cite{IDTA.2023}. The general use-case requires a vendor independent, standard method of describing software functions. Accurate description is crucial for a federated learning approach to enable each partner participation in global model creation. As mentioned earlier, for this research we have used Google AI model cards as a reference\cite{Google.2023}. In addition, we will also implement the updated version of Gaia-X data space connector to provide simultaneous access and download of the service from Gaia-X marketplace to multiple customers \cite{Eclipse.2023}.

\section{Conclusion} 

In this paper, we presented a comprehensive investigation into the effectiveness of a global FedOD model for quality inspection use case in a shared production environment. 
By combining the power of FL and object detection, we achieved substantial improvements in accuracy and precision compared to the individual client models.
Our experimental results showcased the remarkable performance of the FedOD model across multiple scenarios.
The FedOD model outperformed the local models by accurately detecting all combinations and generating highly precise bounding boxes. The achieved average precision (AP) and mean average precision (mAP) scores further demonstrated the model's robustness across multiple unseen test data combinations.
Moreover, the simultaneous evaluation of all three models on live video streams provided valuable insights. The FedOD model consistently demonstrated superior performance in detecting objects, even in the presence of different cabin and windshield combinations.
Furthermore, our experiments on images obtained from the demonstrator in the quality inspection module validated the versatility of the FedOD model. Despite significant variations in background and lighting conditions compared to the training dataset, the model showcased exceptional performance in correctly classifying objects and generating precise bounding boxes. This proves that Federated Object detection is efficient has a wider scope in future collaborative work between multiple partners in manufacturing and other scenarios as well.

\bibliographystyle{IEEEtran.bst}
\bibliography{lit}

\begin{thebibliography}{10}
\providecommand{\url}[1]{#1}
\csname url@samestyle\endcsname
\providecommand{\newblock}{\relax}
\providecommand{\bibinfo}[2]{#2}
\providecommand{\BIBentrySTDinterwordspacing}{\spaceskip=0pt\relax}
\providecommand{\BIBentryALTinterwordstretchfactor}{4}
\providecommand{\BIBentryALTinterwordspacing}{\spaceskip=\fontdimen2\font plus
\BIBentryALTinterwordstretchfactor\fontdimen3\font minus
  \fontdimen4\font\relax}
\providecommand{\BIBforeignlanguage}[2]{{%
\expandafter\ifx\csname l@#1\endcsname\relax
\typeout{** WARNING: IEEEtran.bst: No hyphenation pattern has been}%
\typeout{** loaded for the language `#1'. Using the pattern for}%
\typeout{** the default language instead.}%
\else
\language=\csname l@#1\endcsname
\fi
#2}}
\providecommand{\BIBdecl}{\relax}
\BIBdecl

\bibitem{Redmon.2016}
J.~Redmon, S.~Divvala, R.~Girshick, and A.~Farhadi, ``You only look once:
  Unified, real-time object detection,'' in \emph{Proceedings of the IEEE
  Conference on Computer Vision and Pattern Recognition (CVPR)}, 2016.

\bibitem{Liu.2016}
W.~Liu, D.~Anguelov, D.~Erhan, C.~Szegedy, S.~Reed, C.-Y. Fu, and A.~C. Berg,
  ``Ssd: Single shot multibox detector,'' in \emph{Computer vision - ECCV
  2016}, ser. Lecture Notes in Computer Science, B.~Leibe, J.~Matas, N.~Sebe,
  and M.~Welling, Eds.\hskip 1em plus 0.5em minus 0.4em\relax Cham: Springer,
  2016, vol. 9905, pp. 21--37.

\bibitem{Ren.2016}
S.~Ren, K.~He, R.~Girshick, and J.~Sun, ``Faster r-cnn: Towards real-time
  object detection with region proposal networks,'' 2016.

\bibitem{He.2017}
K.~He, G.~Gkioxari, P.~Dollár, and R.~Girshick, ``Mask r-cnn,'' in \emph{2017
  IEEE International Conference on Computer Vision (ICCV)}, 2017, pp.
  2980--2988.

\bibitem{Tan.2020}
M.~Tan, R.~Pang, and Q.~V. Le, ``Efficientdet: Scalable and efficient object
  detection,'' in \emph{Proceedings of the IEEE/CVF Conference on Computer
  Vision and Pattern Recognition (CVPR)}, June 2020.

\bibitem{McMahan.2017}
B.~McMahan, E.~Moore, D.~Ramage, S.~Hampson, and B.~A.~y. Arcas,
  ``Communication-efficient learning of deep networks from decentralized
  data,'' in \emph{Proceedings of the 20th International Conference on
  Artificial Intelligence and Statistics}, ser. Proceedings of Machine Learning
  Research, A.~Singh and J.~Zhu, Eds., vol.~54.\hskip 1em plus 0.5em minus
  0.4em\relax Fort Lauderdale, FL, USA: PMLR, 2017, pp. 1273--1282.

\bibitem{Hegiste.2022}
V.~Hegiste, T.~Legler, and M.~Ruskowski, ``Application of federated machine
  learning in manufacturing,'' in \emph{2022 International Conference on
  Industry 4.0 Technology (I4Tech)}.\hskip 1em plus 0.5em minus 0.4em\relax
  IEEE, 2022, pp. 1--8.

\bibitem{Jocher.2020}
G.~J. et~al., ``{ultralytics/yolov5: v3.1 - Bug Fixes and Performance
  Improvements},'' Oct. 2020.

\bibitem{Wang.2020}
C.-Y. Wang, A.~Bochkovskiy, and H.-Y.~M. Liao, ``Yolov7: Trainable
  bag-of-freebies sets new state-of-the-art for real-time object detectors,''
  2022.

\bibitem{Luo.2021}
J.~Luo, X.~Wu, Y.~Luo, A.~Huang, Y.~Huang, Y.~Liu, and Q.~Yang, ``Real-world
  image datasets for federated learning,'' 2021.

\bibitem{Bommel.2021}
J.~R. van {Bommel}, ``Active learning during federated learning for object
  detection,'' July 2021.

\bibitem{Zhang.2023}
J.~Zhang, J.~Zhou, J.~Guo, and X.~Sun, ``Visual object detection for
  privacy-preserving federated learning,'' \emph{IEEE Access}, vol.~11, pp.
  33\,324--33\,335, 2023.

\bibitem{He.2021}
C.~He, A.~Shah, Z.~Tang, D.~Fan, A.~N. Sivashunmugam, K.~Bhogaraju, M.~Shimpi,
  L.~Shen, X.~Chu, M.~Soltanolkotabi, and S.~Avestimehr, ``Fedcv: A federated
  learning framework for diverse computer vision tasks,'' \emph{ArXiv}, vol.
  abs/2111.11066, 2021.

\bibitem{Su.2022}
S.~Su, B.~Li, C.~Zhang, M.~Yang, and X.~Xue, ``Cross-domain federated object
  detection,'' 2022.

\bibitem{Lin.2017}
T.-Y. Lin, P.~Goyal, R.~Girshick, K.~He, and P.~Dollár, ``Focal loss for dense
  object detection,'' in \emph{2017 IEEE International Conference on Computer
  Vision (ICCV)}, 2017, pp. 2999--3007.

\bibitem{HeKaiming.2016}
K.~He, X.~Zhang, S.~Ren, and J.~Sun, ``Deep residual learning for image
  recognition,'' in \emph{2016 IEEE Conference on Computer Vision and Pattern
  Recognition (CVPR)}, 2016, pp. 770--778.

\bibitem{Tasnim.2022}
T.~A. Abdel-Aty, E.~Negri, and S.~Galparoli, ``Asset administration shell in
  manufacturing: Applications and relationship with digital twin,''
  \emph{IFAC-PapersOnLine}, vol.~55, no.~10, pp. 2533--2538, 2022, 10th IFAC
  Conference on Manufacturing Modelling, Management and Control MIM 2022.

\bibitem{Alexopoulos.2023}
K.~Alexopoulos, M.~Weber, T.~Trautner, M.~Manns, N.~Nikolakis, M.~Weigold, and
  B.~Engel, ``An industrial data-spaces framework for resilient manufacturing
  value chains,'' \emph{Procedia CIRP}, vol. 116, pp. 299--304, 2023, 30th CIRP
  Life Cycle Engineering Conference.

\bibitem{Jungbluth.2023}
S.~Jungbluth, A.~Witton, J.~Hermann, and M.~Ruskowski, ``Architecture for
  shared production leveraging asset administration shell and gaia-x (in
  press),'' 2023.

\bibitem{Simon.2023}
M.~e.~a. Simon, ``Realisierung einer shared production,'' 2023.

\bibitem{Volkmann.2023}
M.~Volkmann, A.~Wagner, J.~Hermann, and M.~Ruskowski, ``Asset administration
  shells and gaia-x enabled shared production scenario (in press),'' in
  \emph{Lecture Notes in Mechanical Engineering (LNME)}, 2023.

\bibitem{IDTA.2023}
\BIBentryALTinterwordspacing
``Idta registered aas submodel templates,'' 2023. [Online]. Available:
  \url{https://industrialdigitaltwin.org/en/content-hub/submodels}
\BIBentrySTDinterwordspacing

\bibitem{Google.2023}
\BIBentryALTinterwordspacing
``Google cloud model cards,'' 2023. [Online]. Available:
  \url{https://modelcards.withgoogle.com/}
\BIBentrySTDinterwordspacing

\bibitem{Legler.2023}
T.~Legler, V.~Hegiste, and M.~Ruskowski, ``Mapping of newcomer clients in
  federated learning based on activation strength,'' 2023, 32nd International
  Conference Flexible Automation and Intelligent Manufacturing, in press.

\bibitem{Eclipse.2023}
\BIBentryALTinterwordspacing
``Eclipse dataspace components,'' 2023. [Online]. Available:
  \url{https://projects.eclipse.org/projects/technology.edc}
\BIBentrySTDinterwordspacing

\end{thebibliography}


\end{document}